\documentclass{configs/lni}

\usepackage[utf8]{inputenc}
\usepackage{graphicx}
\usepackage{times}
\usepackage{verbatim}
\usepackage{subcaption}
\usepackage{fancyhdr}
\usepackage{amsmath,amssymb,amsfonts}
\usepackage{multirow}
\usepackage{listings} %if lstlistings is used
\usepackage{changepage} %for changing topmargin on first page
\fancypagestyle{titlepage}{
\fancyhead[RO]{\small A. Br\"omme, N. Damer, A. Dantcheva, M. Gomez-Barrero, K. Raja, \linebreak C. Rathgeb, A. Sequeira, and A. Uhl (Eds.): BIOSIG 2022, \linebreak Lecture Notes in Informatics (LNI), Gesellschaft f\"ur Informatik, Bonn 2022} % do NOT modify these lines
\fancyfoot{}}

\renewcommand{\headrulewidth}{0.4pt} %horizontal line below header
\author{Marcel Grimmer \footnote{NTNU, Norwegian Biometrics Laboratory, Gjøvik, marceg@ntnu.no} \, Haoyu Zhang \footnote{NTNU, Norwegian Biometrics Laboratory, Gjøvik, haoyu.zhang@ntnu.no} \,
Raghavendra Ramachandra\footnote{NTNU, Norwegian Biometrics Laboratory, Gjøvik, raghavendra.ramachandra@ntnu.no} \,
Kiran Raja\footnote{NTNU, Norwegian Biometrics Laboratory, Gjøvik, kiran.raja@ntnu.no} \,
Christoph Busch\footnote{h-da, Biometric and Internet-Security Research Group, Darmstadt, christoph.busch@h-da.de} 
}

\title{Time flies by: Analyzing the Impact of Face Ageing on the Recognition Performance with Synthetic Data}

\usepackage[compact]{titlesec}
\titlespacing{\section}{2pt}{*0}{*0}
\titlespacing{\subsection}{2pt}{*0}{*0}
\titlespacing{\subsubsection}{2pt}{*0}{*0}
\let\subparagraph\llncssubparagraph
\usepackage[font=normal,skip=0pt]{caption}
\usepackage{enumitem}
\setlist[itemize]{leftmargin=*}
\let\OLDthebibliography\thebibliography
\renewcommand\thebibliography[1]{
  \OLDthebibliography{#1}
  \setlength{\parskip}{0pt}
  \setlength{\itemsep}{0pt plus 0.3ex}
}
\begin{document}

\maketitle

\renewcommand{\refname}{References}
\setcounter{footnote}{5} %Change to the number of authors for a correct numbering of the foot notes
\thispagestyle{titlepage}
%header setting after the second page
\pagestyle{fancy}
\fancyhead{} % clears header settings
\fancyhead[RO]{\small Analyzing the Impact of Face Ageing on the Recognition Performance with Synthetic Data 
\hspace{25pt}  \hspace{0.05cm}}
\fancyhead[LE]{\hspace{0.05cm}\small  \hspace{25pt} Author 1 and Author 2}
\fancyfoot{} % clears all footer settings
\fancyfoot[LE,RO]{\thepage}
\renewcommand{\headrulewidth}{0.4pt} %line below header

\begin{abstract}
The vast progress in synthetic image synthesis enables the generation of facial images in high resolution and photorealism. In biometric applications, the main motivation for using synthetic data is to solve the shortage of publicly-available biometric data while reducing privacy risks when processing such sensitive information. These advantages are exploited in this work by simulating human face ageing with recent face age modification algorithms to generate mated samples, thereby studying the impact of ageing on the performance of an open-source biometric recognition system. Further, a real dataset is used to evaluate the effects of short-term ageing, comparing the biometric performance to the synthetic domain. The main findings indicate that short-term ageing in the range of 1-5 years has only minor effects on the general recognition performance. However, the correct verification of mated faces with long-term age differences beyond 20 years poses still a significant challenge and requires further investigation.
\end{abstract}
\section{Introduction}
\label{sec:Intro}
The deployment of face recognition systems has gained popularity in various application scenarios, such as border control initiatives like the European Entry-Exit System (EES) \cite{EU-ImplementingDecision-2019-329-on-EES-SampleQuality-190225}. In particular, the EES will be used as a central system for collecting and querying traveller data to the Schengen area at all border crossing points to facilitate the cooperation of visa and law enforcement authorities. The biometric performance of a system deployed in such sensitive environments must comply with high standards, such as those defined in the best practices for automated border control of the European Border and Coast Guard Agency (Frontex) \cite{FRONTEX-BorderControl-BestPractices-InternalDocument-2015}. At the same time, the European General Data Protection Law complicates the processing of biometric data to avoid privacy leakages. \\
Without an appropriate performance testing strategy, the risk of security lapses increases significantly and allows for the discriminatory treatment of travellers due to algorithmic or dataset bias. One solution to the lack of available test data includes the generation of synthetic data samples. However, in order to conduct reliable biometric performance tests, the synthetic samples must be as similar as possible to data collected in operational environments.\\
In the context of synthetic face images, the main focus of this work is to analyse the impact of human face ageing on biometric recognition performance. Due to the 10-year validity of EU passports and enrolment records in immigration systems, face recognition engines employed at the EU borders are frequently exposed to mated face comparisons captured over long time spans. This work deepens the understanding of recognition accuracy and face ageing by analysing synthetically generated face images rendered with ageing effects. This work relies on face age modification methods to avoid the time-consuming data collection of mated samples over time. \\
This analysis is based on face age manipulation frameworks operating within the latent space of StyleGAN \cite{karras2019style} and StyleGAN2 \cite{karras2020analyzing}: InterFaceGAN \cite{shen2020interfacegan} and SAM \cite{alaluf2021matter}. The choice of these techniques is motivated by the high realism and resolution (1024x1024) of facial images the StyleGAN generator achieves. The age-modified face images are analysed with two different face quality assessment algorithms (FQAAs): FaceQnet v1 \cite{hernandez2020biometric} and SER-FIQ \cite{terhorst2020ser}. The biometric performance is further evaluated by computing mated and non-mated comparison scores with ArcFace \cite{deng2019arcface}. The breakdown of mated comparison scores into age bins enables precise testing of the weaknesses of existing face recognition engines. Further, the \textit{UNCW face ageing dataset} (also \textit{MORPH-II}) \cite{ricanek2006morph} is used as a reference for comparing short-term ageing effects to those ageing effects achieved in the synthetic domain. \\
This work is structured as follows: A brief introduction of the face age modification frameworks used to generate the synthetic datasets is given in Section 2. The characteristics of the synthetic and reference datasets are described in Section 3. Finally, the experimental results are presented in Section 4, analysing the FAM ageing accuracy, the biometric quality and comparison scores of synthetic and bona fide data. 
\section{Face Age Modification}
\label{sec:FAP}
This section introduces the basic terms and methods used to create the synthetic cross-age datasets analysed in this work. Face age progression (FAP) refers to rendering from a given input image a synthetic face image with ageing effects, while face age regression (FAR) corresponds to the prediction of rejuvenation effects \cite{grimmer2021deep}. Typically, recent face age modification (FAM) methods predict the appearance of an individual based on a given target age. Another type of FAM technique focuses on changing the age of subjects on a continuous scale with the motivation to better approximate the nature of human ageing. This work evaluates the impact of face ageing on a face recognition (FR) system, using two state-of-the-art FAM frameworks: SAM \cite{alaluf2021matter} and InterFaceGAN \cite{shen2020interfacegan}. \\
Both FAM frameworks are based on manipulating latent vectors in the latent space of StyleGAN \cite{karras2019style} and StyleGAN2 \cite{karras2020analyzing}. The main idea is to exploit the disentanglement of facial attributes given in the internal data representation of a generative adversarial network (GAN). Operating directly in the latent space of a pre-trained GAN alleviates the need to train complex adversarial networks and benefits from the high resolution and photorealism achieved by the StyleGAN generators.

 \begin{figure}
		\centering \includegraphics[scale=0.4]{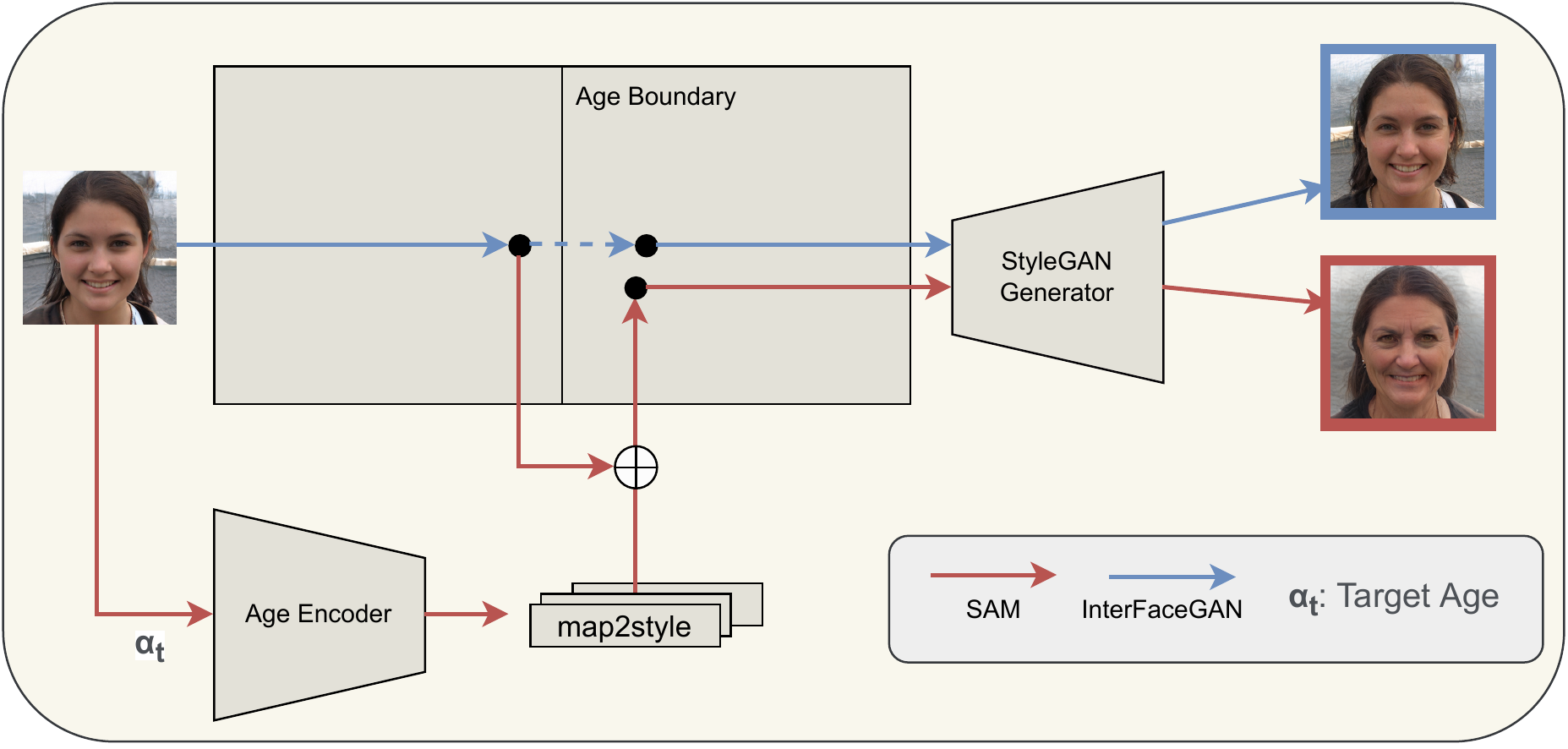}
		\caption{Latent face ageing with SAM and InterFaceGAN}
		\label{fig:fam-overview}
\end{figure}

The basic FAM principles of SAM and InterFaceGAN are illustrated in Figure \ref{fig:fam-overview}. The main question is where to move the randomly drawn latent vector to change the age while leaving other facial attributes unchanged \cite{grimmer2021generation}. InterFaceGAN addresses this issue by training a binary age boundary that divides the latent space into two subspaces (old vs young). Afterwards, the age is increased by moving an arbitrary latent vector into the perpendicular direction of the age boundary, with the magnitude defining the ageing extent. \\
Unlike InterFaceGAN, SAM trains an additional age encoder conditioned on the target age $\alpha_{t}$, extracting the missing ageing patterns by learning the residuals to the original face image. In a next step, a pre-trained map2style network \cite{richardson2021encoding} transforms the residual ageing patterns into latent codes, which are then fused with the initial latent vector randomly drawn from the latent space. After fusing the residual age patterns with the initial latent vector, the resulting latent code is passed to the StyleGAN2 generator to generate the age-modified face image.

\section{Datasets}
\label{sec:Datasets}
This section introduces the synthetic cross-age datasets generated with InterFaceGAN and SAM, as well as the bona fide reference datasets (FRGC v2.0 \cite{FRGC_DB}, UNCW ageing dataset \cite{ricanek2006morph}).

\subsection{Synthetic Dataset Generation}

Our base synthetic data is randomly generated by the StyleGAN \cite{karras2019style} generator pre-trained on the FFHQ \cite{karras2019style} dataset. Choosing a truncation factor of $\psi = 0.75$ has proven as an effective setup \cite{zhang2021applicability} for generating visually appealing face images with a high diversity of demographic factors. Further, the work of Zhang et al. \cite{zhang2021applicability} indicates minor differences in the face recognition performance between StyleGAN and StyleGAN2 generated face images. Therefore, we select a dataset of $50,000$ face images generated with StyleGAN as a basis for our face age modification algorithms. \\
Given these synthetic base images, the corresponding age-modified samples are generated using the proposed semantic editing algorithms of Shen et al.\cite{shen2020interfacegan} (InterFaceGAN) and Alaluf et al.\cite{alaluf2021matter}. As shown in Figure \ref{fig:fam-overview}, InterFaceGAN controls the shifting distance in the latent space with a scaling factor that we empirically set as $s_1\pm 0.4, s_2 \pm 0.8, s_3 \pm 1.2$ to create 6 synthetic data subsets containing mated samples of the base images. \\
As introduced in Section \ref{sec:FAP}, the input of SAM \cite{alaluf2021matter} is a target age and a base image. We select 7 different target age groups (10, 20, 30, 40, 50, 60, 70) to which we transform the base synthetic images to. In the original SAM algorithm, the pixel2style2pixel (pSp) \cite{richardson2021encoding} encoder is applied to first project the base images into the extended StyleGAN2 latent space ($W+$) in order to fuse it with the encoded age residual code. In this work, we discard the initial base image as soon as it is projected to the $W+$ latent space, re-defining the reconstructed face image as our new base in order to avoid the distortion of our results due to identity losses caused by GAN inversion. \\
In this work, face images with unrealistic capturing conditions are filtered out to increase the representativeness of our datasets. Details of the filtering pipeline are given in Table \ref{tab:dataset_general}. For the inter-eye-distance (IED), a pre-trained landmark detection model is used to predict the centre of the eyes and filter out images with IED less than 90 pixels or failed landmark detections. To filter out images with unsatisfying illumination conditions, we prepare an internal dataset with binary labels (good or poor illumination condition) and train a random forest regressor on the extracted features that measure illumination uniformity and symmetry from these images. The Img2pose model \cite{albiero2021img2pose} is applied to predict the Euler angles of the head pose and filter out images with extreme rotations. Additionally, we included C3AE \cite{zhang2019c3ae} to predict the age of the given images and exclude those with ages not in the range of [13,59] years. Finally, Table \ref{tab:dataset_general} illustrates the exact number of face images in the analysed datasets - before and after applying the filtering pipeline.

% \usepackage{multirow}
%\begin{table}[]
%\centering
%\begin{tabular}{|c|c|}
%\hline
%Filter                     & Unqualified condition(s)                  \\ \hline
%Inter-eye-distance (IED)   & IED $\leq$ 90                                     \\ \hline
%Illumination condition     & Classification result of pretrained model \\ \hline
%\multirow{3}{*}{Head pose} & Yaw angle $\notin$ [-15°, 15°]                               \\ %\cline{2-2} 
%                           & or pitch angle $\notin$ [ -25°, 20°]                           \\ %\cline{2-2} 
%                           & or roll angle $\notin$ [-20°, 20°]                            \\ %\hline
%Age                        & Estimated age $\notin$ [13, 59] years old                  \\ \hline
%\end{tabular}
%\label{tab:dataset_filter}
%\caption{Overview of the filtering criteria}
%\end{table}

\begin{table}[]
\centering
\resizebox{.6\linewidth}{!}{
\begin{tabular}{|c|c|c|}
\hline
Dataset                                   & \#Images before Filtering & \#Images after Filtering \\ \hline
FRGC v2                                   & $24,025$                    & $17,919$                   \\ \hline
Synthetic Base                            & $50,000$                    & $25,918$                   \\ \hline
InterFaceGAN (scale = 0.4)                & $51,836$                    & $48,513$                   \\ \hline
InterFaceGAN (scale = 0.8)                & $51,836$                    & $47,085$                   \\ \hline
InterFaceGAN (scale = 1.2)                & $51,836$                    & $44,858$                   \\ \hline
SAM (target age = 10)                     & $25,918$                    & $18,290$                   \\ \hline
SAM (target age = 20)                     & $25,918$                    & $22,671$                   \\ \hline
SAM (target age = 30)                     & $25,918$                    & $23,253$                   \\ \hline
SAM (target age = 40)                     & $25,918$                    & $23,513$                   \\ \hline
SAM (target age = 50)                     & $25,918$                    & $22,671$                   \\ \hline
SAM (target age = 60)                     & $25,918$                    & $17,174$                   \\ \hline
SAM (target age = 70)                     & $25,918$                    & $10,028$                   \\ \hline
\end{tabular}
}
\caption{General Database Information \label{tab:dataset_general}}
\end{table}

\subsection{Bona Fide Reference Datasets}
\label{sec:synth-vs-real-datasets}
To compare the synthetic data with real data, we choose a representative dataset containing $17,919$ images from FRGC-V2 \cite{FRGC_DB}, which is known for its high-quality images and constrained conditions resembling those of border crossing capturing environments. However, despite the good representativeness, FRGC-V2 samples are not annotated with ground-truth ages, thus limiting the age-based performance comparison to the synthetic datasets. \\
To overcome this limitation, the UNCW face ageing dataset \cite{ricanek2006morph} is further used in our analysis, including more than $55,000$ face images of more than $13,000$ individuals with exact age annotations, where for mated comparison trials the difference in age is ranging from 164 days to $1,681$ days. In order to analyse short-term ageing effects and their impact on the face recognition performance, we sorted out mated samples with less than 1 year passed between the probe and reference image capturing, leaving an amount of $37,423$ face images. Similarly, the synthetic datasets have been further reduced to only include mated pairs with age differences within 1 to 5 years. As no age labels are given for the synthetic data, we apply the C3AE \cite{zhang2019c3ae} age estimator to predict the age labels for each face image individually. The resulting short-term InterFaceGAN (ST-InterFaceGAN) and SAM (ST-SAM) datasets comprise $10,772$ and $12,298$ samples.

%\begin{table}[]
%\centering
%\resizebox{.5\linewidth}{!}{
%\begin{tabular}{|c|c|c|}
%\hline
%Dataset                        & \#Images & Age Differences (years) \\ \hline
%FRGC v2                        & 17,919   & -                                    \\ \hline
%ST-UNCW       & 37,423   & 1-5                            \\ \hline
%ST-InterFaceGAN & 10,772   & 1-5                            \\ \hline
%ST-SAM                     & 12,298   & 1-5                            \\ \hline
%\end{tabular}
%}
%\caption{Information of short-term (ST) ageing database \label{tab:dataset_agediff}}
%\end{table}
\section{Experimental Results}
\label{sec:Exp}

%%%%%%%%%%%%%%%%%%%%%% Only arxiv version
\subsection{Ageing Accuracy}
\label{sec:exp_accuracy}
\begin{figure}
\centering
\resizebox{.9\linewidth}{!}{
  \begin{subfigure}[b]{0.48\columnwidth}
    \includegraphics[width=\linewidth]{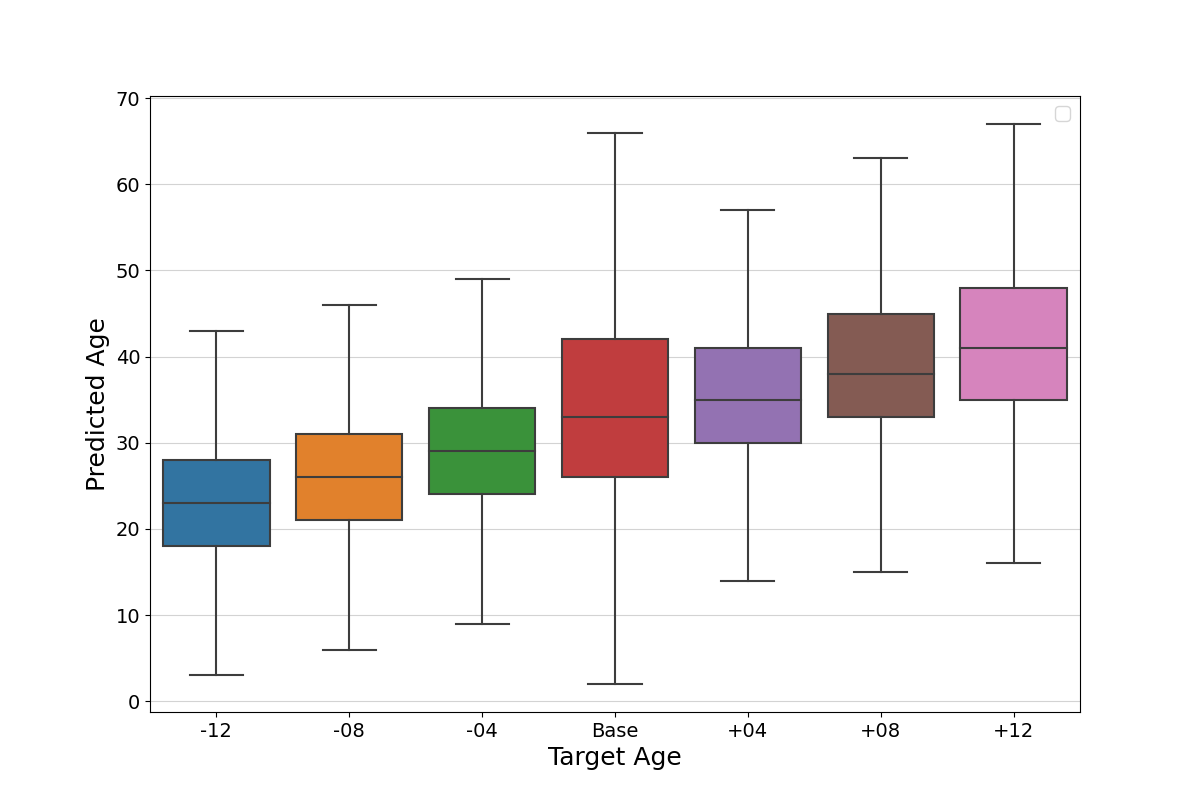}
    %\caption{InterFaceGAN with different editing scales}
    %\label{fig:age_accuracy_InterFaceGAN}
  \end{subfigure}
  \begin{subfigure}[b]{0.48\columnwidth}
    \includegraphics[width=\linewidth]{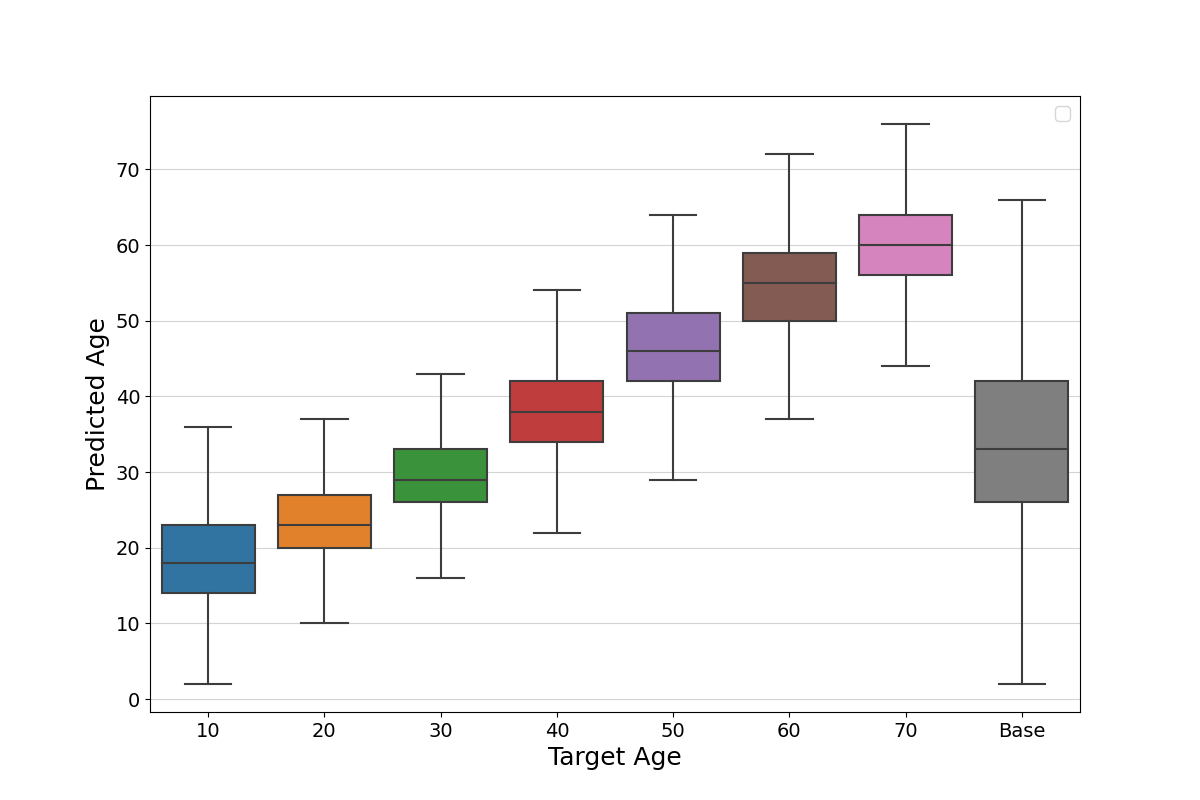}    
    %\caption{SAM for different targe ages}
    %\label{fig:age_accuracy_SAM}
  \end{subfigure}
}
  \caption{Ageing accuracy analysis of InterFaceGAN (left) and SAM (right) \label{fig:age_accuracy} }
\end{figure}

This section evaluates the effectiveness of the FAM frameworks in terms of their ageing accuracy. Since the analysed face images are fully synthetic, no ground-truth age labels are given to assess the exact age difference between the target and the actual age of the individuals. Instead, a pre-trained age estimation network (C3AE) \cite{zhang2019c3ae} is utilised to predict the ground-truth age labels to determine the ageing accuracy performance. The predicted ages of face images generated with InterFaceGAN and SAM are visualised as boxplots in Figure \ref{fig:age_accuracy}. 

By the nature of InterFaceGAN, it is not possible in the interaction with the latent vectors to specify target ages since ageing is achieved by continuously shifting the latent vectors in the non-linear latent space of StyleGAN. Therefore, Figure \ref{fig:age_accuracy} aims to strengthen the understanding between latent space distances and their corresponding age differences. While the red boxplot shows the predicted ages of the base images, the decreasing medians of the left-sided boxplots verify the effectiveness of InterFaceGAN for FAR. Meanwhile, the increasing trend of medians of the right-sided boxplots confirms the same effectiveness for FAP. However, it is also noticeable that the same distances can lead to a diversity of age differences thus demonstrating the non-linearity of the latent space.

Additionally, Figure \ref{fig:age_accuracy} presents the ageing accuracy results of SAM, comparing the target ages (x-axis) with the predicted ages (y-axis). Other than InterFaceGAN, SAM enables us to specify target ages, thus allowing to evaluate the ageing accuracy performance more precisely. That to say, a linear increase of the boxplot’s medians (horizontal lines within boxes) corresponds to a perfect ageing accuracy since the average predicted age equals the target ages. By analysing and comparing the medians, SAM proves to reliably manipulate ages with target ages chosen between 20 and 40 years. However, the ageing accuracy of SAM drops for more extreme target ages – i.e., children and seniors. This observation is a typical sign of biased image generation models, which were trained on unbalanced datasets with unequal distribution of soft biometrics. Since SAM utilises the pre-trained generator of StyleGAN2 trained on the web-crawled FFHQ dataset, most of the individuals seen during training are within 30 to 40 years. As a consequence, the generator fails to learn the craniofacial growth occurring during adolescence, as well as the intensifying of wrinkles and furrows occurring with older ages. The issue of demographic biases is due to a general data scarcity of samples having underrepresented characteristics. Despite the so caused inaccuracies, the average ageing span of SAM includes 41 years, ranging from 19 to 60 years – thus suitable to simulate longer-term ageing patterns. 

%%%%%%%%%%%%%%%%%%%%%%

\subsection{Face Image Quality Assessment}
\label{sec:exp_FIQA}

\begin{figure}
\centering
\resizebox{.8\linewidth}{!}{
  \begin{subfigure}[b]{0.6\columnwidth}
    \includegraphics[width=\linewidth]{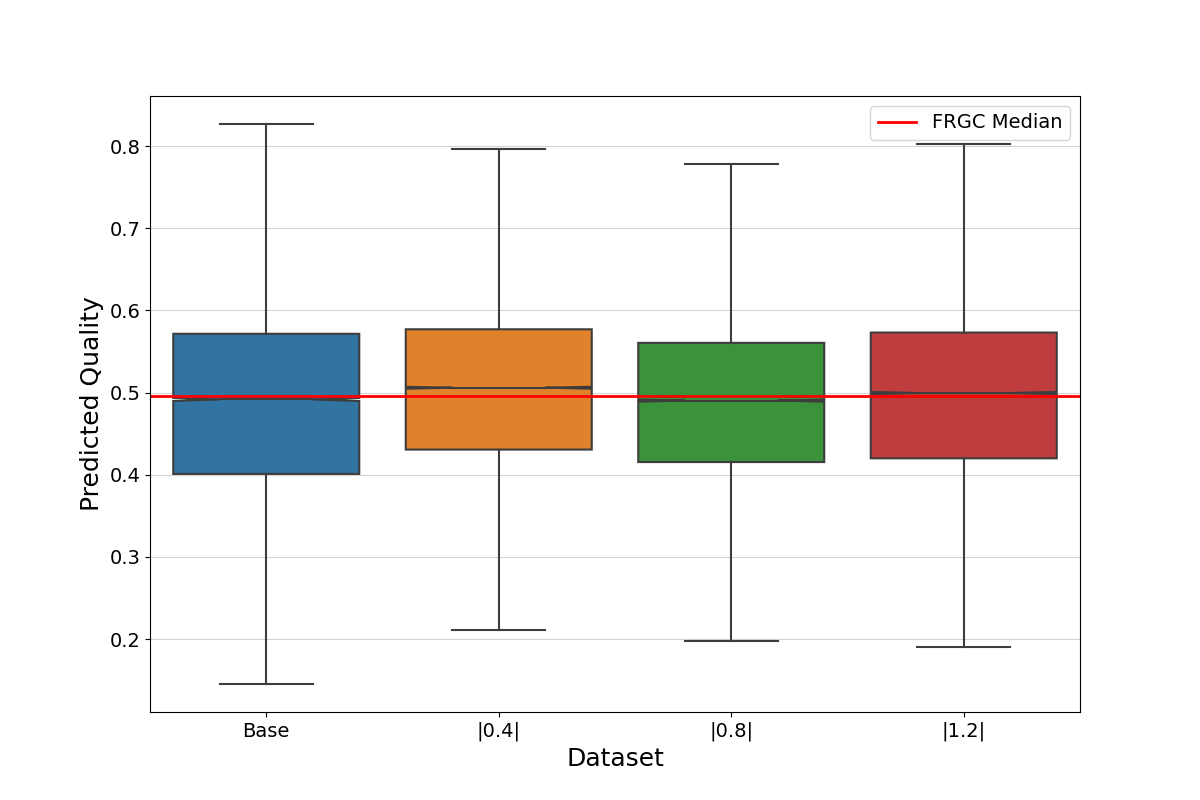}
    %\caption{FaceQnet v1}
    %\label{fig:interfacegan-faceqnetv1}
  \end{subfigure}
  \begin{subfigure}[b]{0.6\columnwidth}
    \includegraphics[width=\linewidth]{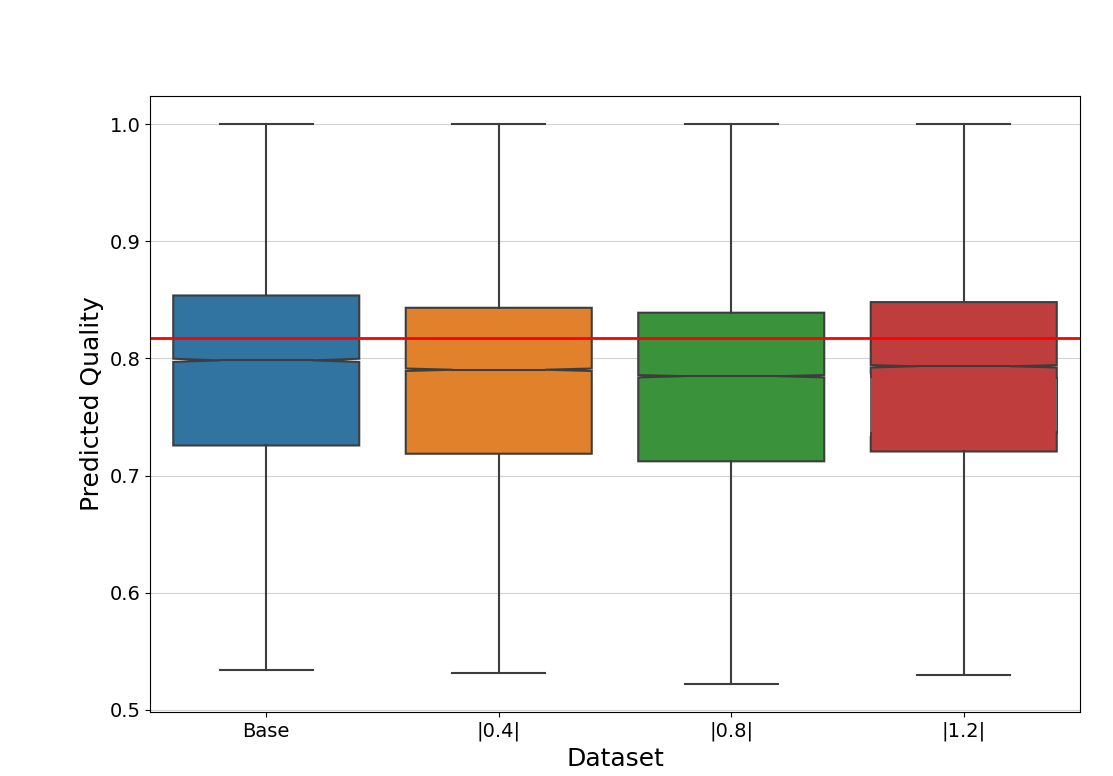}    %\caption{SER-FIQ}
    %\label{fig:interfacegan-serfiq}
  \end{subfigure}
}
 
 \resizebox{.8\linewidth}{!}{
  \begin{subfigure}[b]{0.59\columnwidth}
    \includegraphics[width=\linewidth]{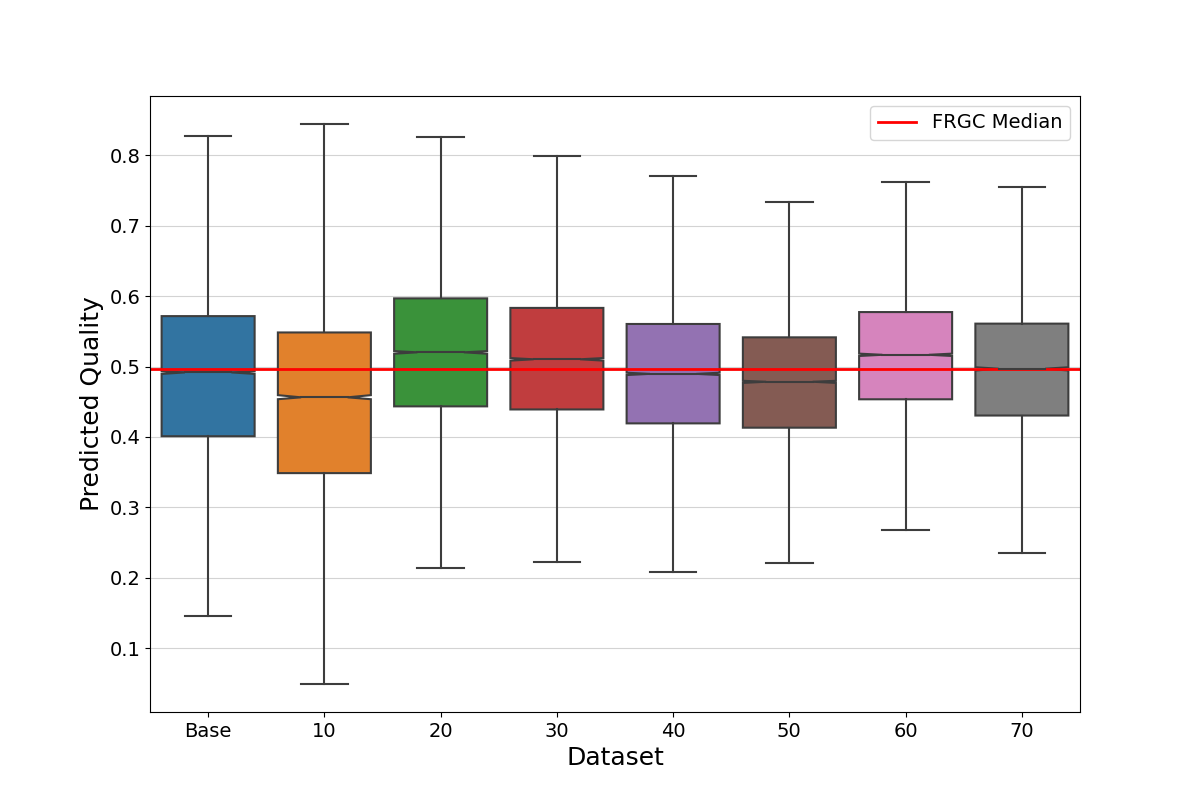}
    %\caption{FaceQnet v1}
    %\label{fig:sam-faceqnetv1}
  \end{subfigure}
   \begin{subfigure}[b]{0.59\columnwidth}
    \includegraphics[width=\linewidth]{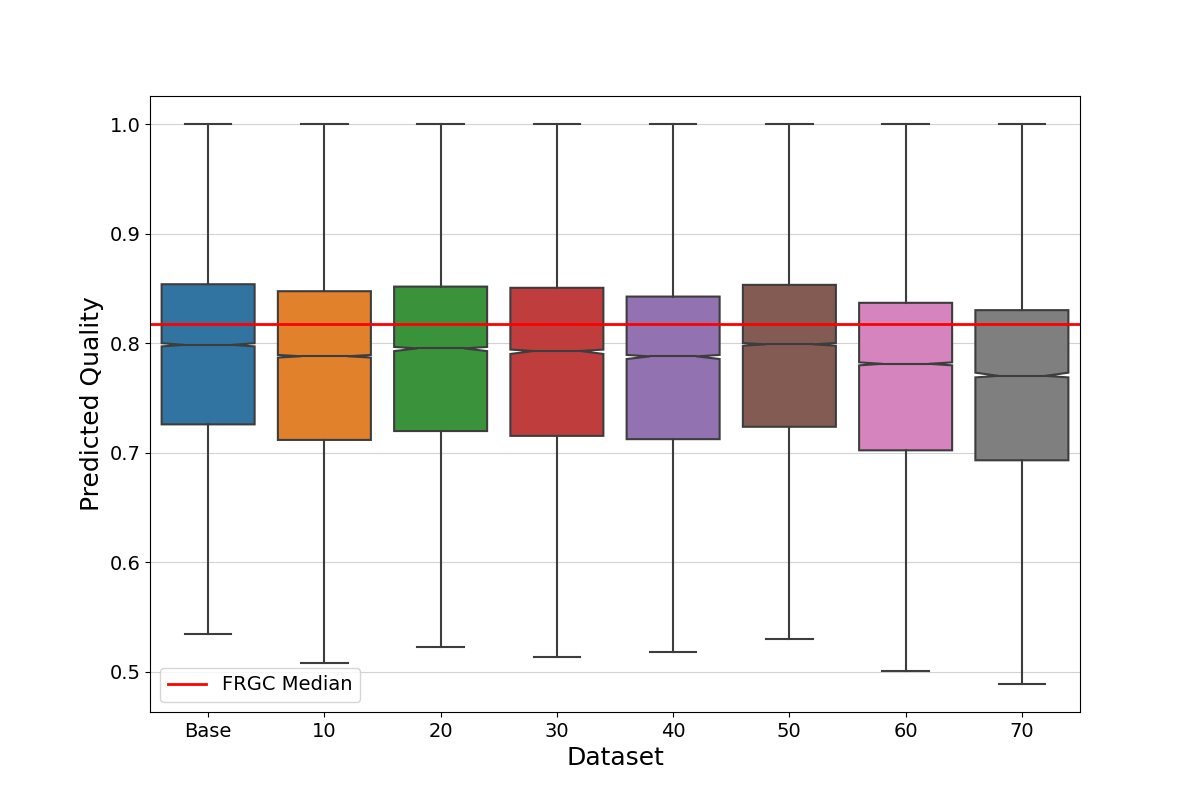}    %\caption{SER-FIQ}
    %\label{fig:sam-serfiq}
  \end{subfigure}
}
  \caption{Biometric quality analysis of synthetic images generated with InterFaceGAN (top) and SAM (bottom). The biometric quality is estimated with two FQAAs: SER-FIQ (right) and FaceQnet v1 (left). The red line visualises the median biometric quality of the bona fide reference dataset (FRGC-V2). \label{fig:boxplots_biometric_quality}}
\end{figure}

The main goal of this subsection is to compare the age-modified datasets with the reference bona fide datasets by utilising face quality assessment algorithms (FQAAs). FQAAs are developed to predict the biometric quality of a given face image by translating its suitability for face recognition to a scalar value between [0, 1] (1= “Perfect biometric face image quality”, 0=”worst biometric quality”). In particular, two well-established deep learning-based FQAAs are selected for this task: FaceQnet v1 \cite{hernandez2020biometric} and SER-FIQ \cite{terhorst2020ser}. \\
In Figure \ref{fig:boxplots_biometric_quality}, the notched boxplots visualise the median face image qualities with their 95\%-confidence intervals. In addition, the horizontal red line represents the median of the FRGC-V2 reference dataset. This view enables the analysis of statistical deviations of the synthetic datasets' medians to the median biometric quality of real data. In this context, Figure \ref{fig:boxplots_biometric_quality} reveals that all boxplots enclose the red line, thus indicating no statistical differences in the biometric quality across all age-modified datasets. \\
However, it is noticeable that the medians of the synthetic datasets estimated with SER-FIQ consistently falls below the red line, thus supporting the conclusion of minor biometric quality differences between synthetic and real data. In contrast, the medians estimated with FaceQnet v1 fluctuate below and and above the red reference line, hence strengthening the "no difference" hypothesis.

% InterfaceGAN

\subsection{Comparison Score Analysis}
\label{sec:exp_cscore}
\begin{figure}
\centering
\resizebox{.8\linewidth}{!}{
  \begin{subfigure}[b]{0.47\columnwidth}
    \includegraphics[width=\linewidth]{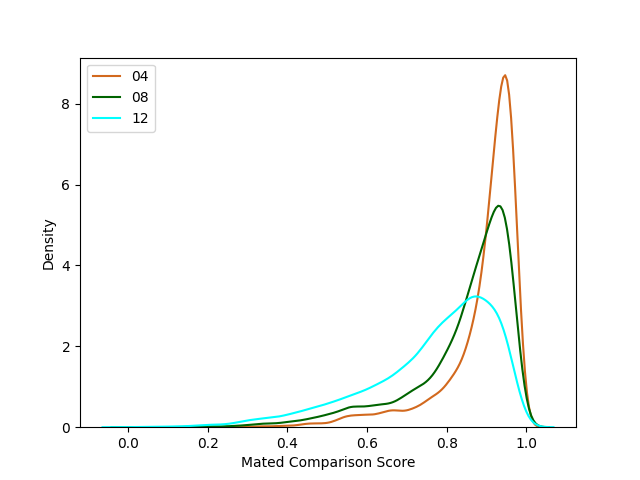}
    %\label{fig:cscore_mated_interfacegan}
  \end{subfigure}
  \begin{subfigure}[b]{0.47\columnwidth}
    \includegraphics[width=\linewidth]{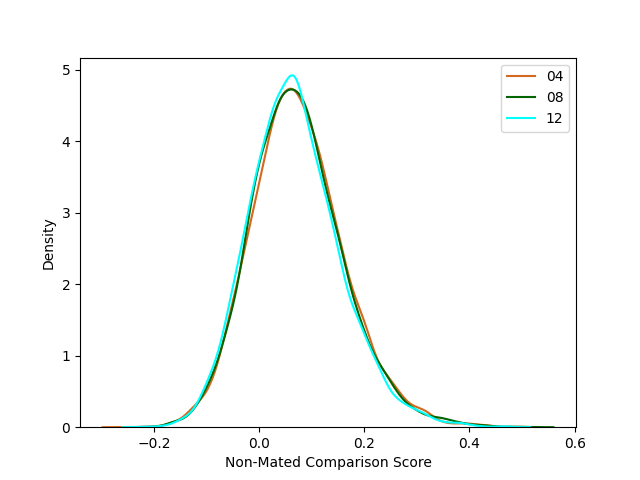}      
    %\label{fig:cscore_nonmated_interfacegan}
  \end{subfigure}
}

\resizebox{.8\linewidth}{!}{  
  \begin{subfigure}[b]{0.47\columnwidth}
    \includegraphics[width=\linewidth]{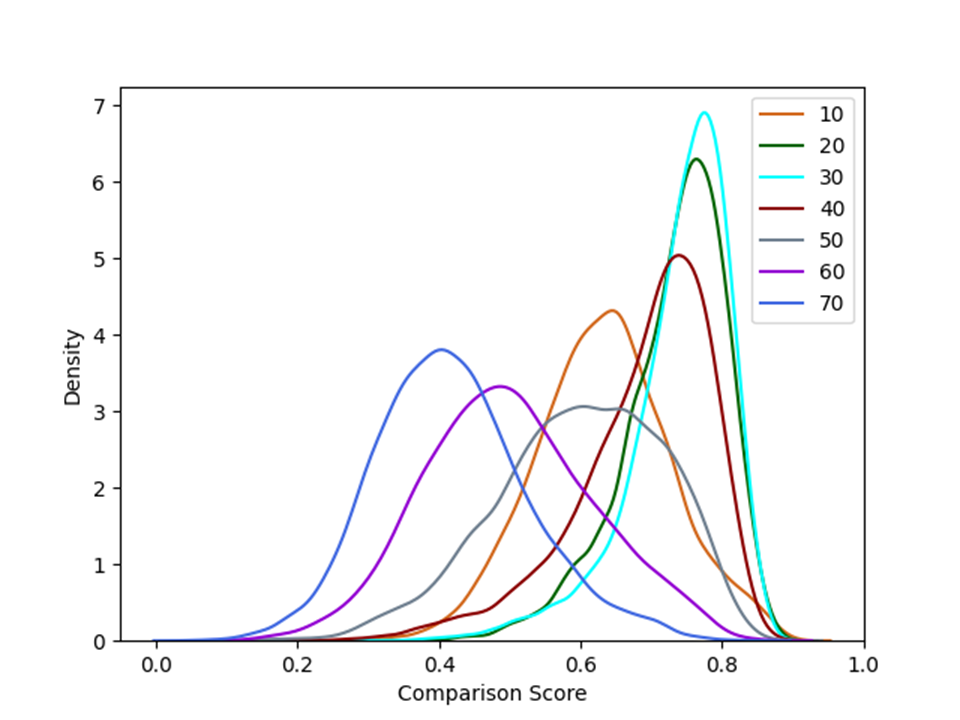}
    %\label{fig:cscore_mated_sam_inv}
  \end{subfigure}
  \begin{subfigure}[b]{0.47\columnwidth}
    \includegraphics[width=\linewidth]{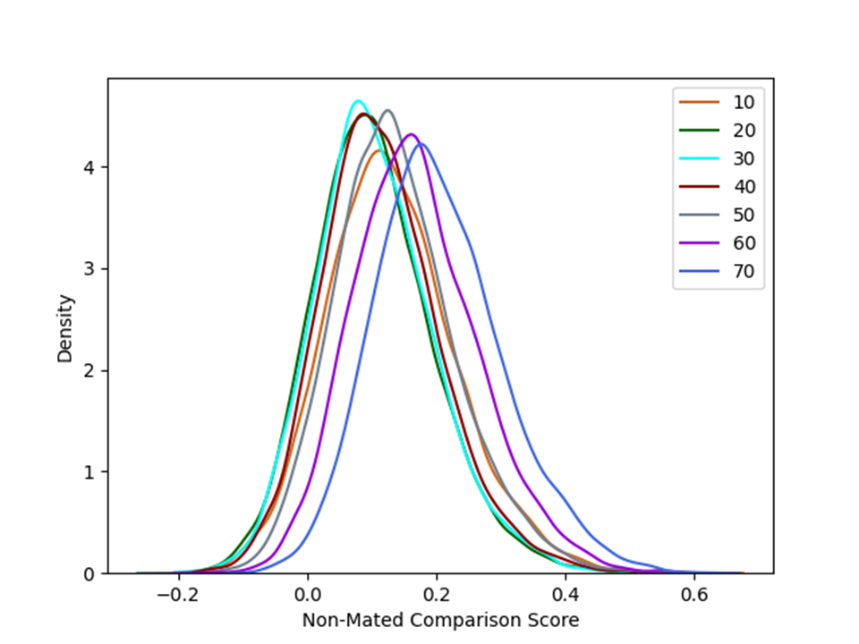}  
    %\label{fig:cscore_mated_sam_inv}
  \end{subfigure}
 }
  \caption{Mated (left column) and non-mated (right column) comparison scores based on age-modified datasets generated with InterFaceGAN (top) and SAM (bottom) \label{fig:cscore_interfacegan_sam}}
\end{figure}

This section analyses the mated and non-mated comparison scores obtained by comparing each base image to the age-modified versions (mated) and other age-modified identities (non-mated). Figure \ref{fig:cscore_interfacegan_sam} shows that increasing age modifications lead to decreasing similarities between the synthetic mated samples generated with InterFaceGAN. This observation corresponds to our initial hypothesis that existing face recognition systems are not trained to compensate ageing effects. The more years pass between the enrolment process and the re-capturing of a probe image, the more intense ageing effects will occur and thus affect the recognition performance. On the contrary, the non-mated comparison scores illustrate only minor performance differences, as seen by the nearly identical distributions. \\
Similarly, Figure \ref{fig:cscore_interfacegan_sam} shows the kernel density plots of comparison scores obtained by comparing synthetic samples generated with SAM. The general behaviour of the mated and non-mated comparison scores is similar to the results reported with InterFaceGAN. The more the target age differs from the average StyleGAN age (34y), the less similar the mated samples become. This loss in identity over time can either be caused by the FAM algorithms or the incapability of ArcFace to handle long-term age differences between the probe and reference sample. That to say, the disentanglement of these two sources of potential identity loss remains a challenging task.

\subsubsection{Synthetic vs Natural Face Ageing}
\label{sec:exp_cscore_agediff}

\begin{figure}
\centering
\resizebox{.8\linewidth}{!}{
  \begin{subfigure}[b]{0.47\columnwidth}
    \includegraphics[width=\linewidth]{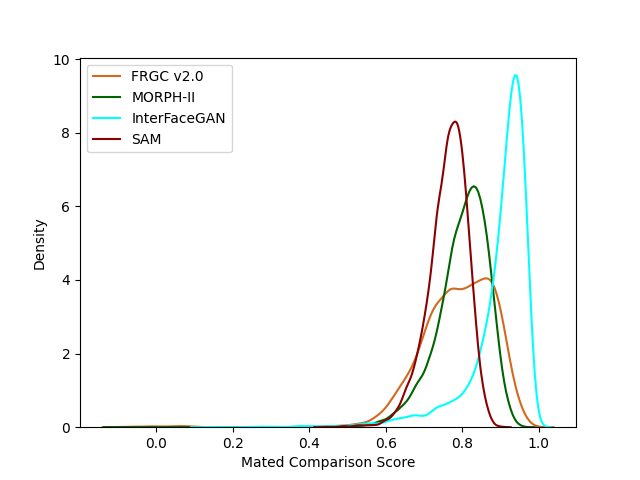}
  \end{subfigure}
  \begin{subfigure}[b]{0.47\columnwidth}
    \includegraphics[width=\linewidth]{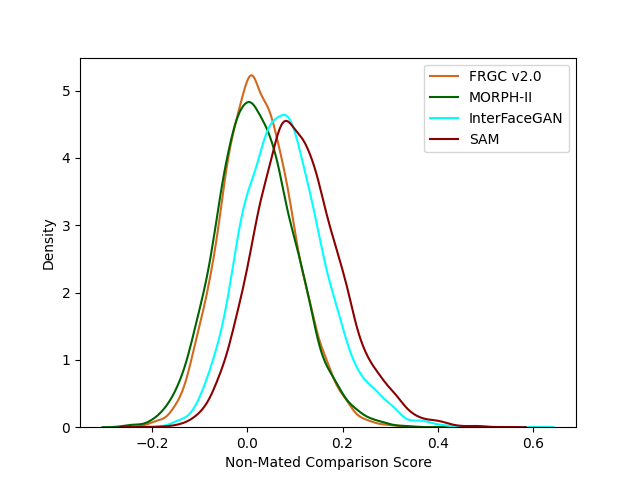}    
  \end{subfigure}
}
  \caption{Mated (left) and non-mated (right) comparison scores based on synthetic and real datasets with age difference within 1-5 years \label{fig:cscore_agediff}}
\end{figure}

The final part of this section analyses the impact of short-term ageing effects based on the ST-ageing datasets introduced in Section \ref{sec:synth-vs-real-datasets}. Figure \ref{fig:cscore_agediff} shows the mated comparison score distributions, indicating no significant differences between mated samples collected without ageing (FRGC v2.0) and those with age gaps  within 1 to 5 years (UNCW). The main reason for this observation is most likely due to minor ageing patterns given the short time intervals available in this analysis. A similar behaviour is observed with SAM generated face images (red line), hence indicating that the synthetic mated samples are similar to those seen in real data. \\
Finally, the cyan curve achieved the highest similarity scores, emphasising the capability of InterFaceGAN to preserve identity information during the age synthesis. Despite the effective identity preservation rate, the comparison to the real curves reveals a large domain gap, thus potentially overestimating mated comparison scores observed in real-world settings. Another domain-gap crystallizes in the non-mated comparison scores in Figure \ref{fig:cscore_agediff}: While the synthetic lines are nearly identical, their average scores are higher than those measured for bona fide data. 

\section{Conclusion}
\label{sec:Conclusion}

The main focus of this work is to analyse the impact of face ageing on face recognition systems by using FAM to generate synthetic mated and non-mated samples with varying age gaps. For this purpose, the FAM performance is analysed in terms of the biometric quality (Section \ref{sec:exp_FIQA}) and identity preservation (Section \ref{sec:exp_cscore}) of the generated mated and non-mated face images. The main findings of this work underline the capability of synthetic face images to interfere with face recognition systems similar than bona fide data. Further, the comparison score analysis indicates only a minor deterioration in the recognition performance for short and medium-term ageing intervals - as shown by comparisons conducted in the synthetic and real domain. Nevertheless, the mated comparison scores significantly decrease for long-term age intervals or extreme target age choices. \\ Finally, this work demonstrates the future value of synthetic face images in analysing the age-robustness of FR systems. Accelerated by the remarkable progress of deep generative networks, we believe the domain gap between synthetic and bona fide data to vanish over time. Especially in the context of face ageing, FAM algorithms are crucial for avoiding long-lasting data collection initiatives, which are not feasible given the time constraints of real-world applications. In an endeavour of closing the domain gap between synthetic and bona fide data, future research may benefit from new concepts, such as 3D multi-view image synthesis \cite{chan2022efficient} in order to better preserve spatial information of the faces and support the generation of geometry-consistent mated samples.

\bibliography{lniguide}

\begin{thebibliography}{APCO21}

\bibitem[Al21]{albiero2021img2pose}
Albiero, Vitor; Chen, Xingyu; Yin, Xi; Pang, Guan; Hassner, Tal: img2pose: Face
  alignment and detection via 6dof, face pose estimation.
\newblock In: Proceedings of the IEEE/CVF Conference on Computer Vision and
  Pattern Recognition.
\newblock pp. 7617--7627, 2021.

\bibitem[APCO21]{alaluf2021matter}
Alaluf, Yuval; Patashnik, Or; Cohen-Or, Daniel: Only a Matter of Style: Age
  Transformation Using a Style-Based Regression Model.
\newblock ACM Trans. Graph., 40(4), 2021.

\bibitem[Ch22]{chan2022efficient}
Chan, Eric~R; Lin, Connor~Z; Chan, Matthew~A; Nagano, Koki; Pan, Boxiao;
  De~Mello, Shalini; Gallo, Orazio; Guibas, Leonidas~J; Tremblay, Jonathan;
  Khamis, Sameh et~al.: Efficient geometry-aware 3D generative adversarial
  networks.
\newblock In: Proceedings of the IEEE/CVF Conference on Computer Vision and
  Pattern Recognition.
\newblock pp. 16123--16133, 2022.

\bibitem[De19]{deng2019arcface}
Deng, Jiankang; Guo, Jia; Xue, Niannan; Zafeiriou, Stefanos: Arcface: Additive
  angular margin loss for deep face recognition.
\newblock In: Proceedings of the IEEE/CVF conference on computer vision and
  pattern recognition.
\newblock pp. 4690--4699, 2019.

\bibitem[{Eu}19]{EU-ImplementingDecision-2019-329-on-EES-SampleQuality-190225}
{European Council}: , Commission Implementing Decision 2019/329 of 25 February
  2019 laying down the specifications for the quality, resolution and use of
  fingerprints and facial image for biometric verification and identification
  in the Entry/Exit System ({EES}), February 2019.

\bibitem[{Fr}15]{FRONTEX-BorderControl-BestPractices-InternalDocument-2015}
{Frontex}: , {Best practice technical guidelines for Automated Border Control
  (ABC) systems}, 2015.

\bibitem[Gr21]{grimmer2021generation}
Grimmer, Marcel; Zhang, Haoyu; Ramachandra, R.; Raja, K.; Busch, C.: Generation
  of Non-Deterministic Synthetic Face Datasets Guided by Identity Priors.
\newblock arXiv preprint arXiv:2112.03632, 2021.

\bibitem[GRB21]{grimmer2021deep}
Grimmer, Marcel; Ramachandra, Raghavendra; Busch, Christoph: Deep face age
  progression: A survey.
\newblock IEEE Access, 9:83376--83393, 2021.

\bibitem[He20]{hernandez2020biometric}
Hernandez-Ortega, Javier; Galbally, Javier; Fierrez, Julian; Beslay, Laurent:
  Biometric quality: Review and application to face recognition with faceqnet.
\newblock arXiv preprint arXiv:2006.03298, 2020.

\bibitem[Ka20]{karras2020analyzing}
Karras, Tero; Laine, Samuli; Aittala, Miika; Hellsten, Janne; Lehtinen, Jaakko;
  Aila, Timo: Analyzing and improving the image quality of stylegan.
\newblock In: Proceedings of the IEEE/CVF Conference on Computer Vision and
  Pattern Recognition.
\newblock pp. 8110--8119, 2020.

\bibitem[KLA19]{karras2019style}
Karras, Tero; Laine, Samuli; Aila, Timo: A style-based generator architecture
  for generative adversarial networks.
\newblock In: Proceedings of the IEEE/CVF Conference on Computer Vision and
  Pattern Recognition.
\newblock pp. 4401--4410, 2019.

\bibitem[{Ph}05]{FRGC_DB}
{Phillips}, P.~J.; {Flynn}, P.~J.; {Scruggs}, T.; {Bowyer}, K.~W.; {Jin Chang};
  {Hoffman}, K.; {Marques}, J.; {Jaesik Min}; {Worek}, W.: Overview of the face
  recognition grand challenge.
\newblock In: 2005 IEEE Computer Society Conference on Computer Vision and
  Pattern Recognition (CVPR'05).
\newblock pp. 947--954 vol. 1, June 2005.

\bibitem[Ri21]{richardson2021encoding}
Richardson, Elad; Alaluf, Yuval; Patashnik, Or; Nitzan, Yotam; Azar, Yaniv;
  Shapiro, Stav; Cohen-Or, Daniel: Encoding in style: a stylegan encoder for
  image-to-image translation.
\newblock In: Proceedings of the IEEE/CVF Conference on Computer Vision and
  Pattern Recognition.
\newblock pp. 2287--2296, 2021.

\bibitem[RT06]{ricanek2006morph}
Ricanek, Karl; Tesafaye, Tamirat: Morph: A longitudinal image database of
  normal adult age-progression.
\newblock In: 7th international conference on automatic face and gesture
  recognition (FGR06).
\newblock IEEE, pp. 341--345, 2006.

\bibitem[Sh20]{shen2020interfacegan}
Shen, Yujun; Yang, Ceyuan; Tang, Xiaoou; Zhou, Bolei: Interfacegan:
  Interpreting the disentangled face representation learned by gans.
\newblock IEEE transactions on pattern analysis and machine intelligence, 2020.

\bibitem[Te20]{terhorst2020ser}
Terhorst, Philipp; Kolf, Jan~Niklas; Damer, Naser; Kirchbuchner, Florian;
  Kuijper, Arjan: SER-FIQ: Unsupervised estimation of face image quality based
  on stochastic embedding robustness.
\newblock In: Proceedings of the IEEE/CVF Conference on Computer Vision and
  Pattern Recognition.
\newblock pp. 5651--5660, 2020.

\bibitem[Zh19]{zhang2019c3ae}
Zhang, Chao; Liu, Shuaicheng; Xu, Xun; Zhu, Ce: C3AE: Exploring the limits of
  compact model for age estimation.
\newblock In: Proceedings of the IEEE/CVF Conference on Computer Vision and
  Pattern Recognition.
\newblock pp. 12587--12596, 2019.

\bibitem[Zh21]{zhang2021applicability}
Zhang, Haoyu; Grimmer, Marcel; Ramachandra, Raghavendra; Raja, Kiran; Busch,
  Christoph: On the Applicability of Synthetic Data for Face Recognition.
\newblock In: 2021 IEEE International Workshop on Biometrics and Forensics
  (IWBF).
\newblock IEEE, pp. 1--6, 2021.

\end{thebibliography}

\end{document}